%% file: main.tex
\documentclass{article}

\usepackage[T1]{fontenc}
\usepackage[english]{babel}
\usepackage{csquotes}

\usepackage{hyperref}
\usepackage{mathtools}
\usepackage{graphicx}
\usepackage{algorithm}
\usepackage{algpseudocode}
\usepackage{amsmath}
\usepackage[capitalize]{cleveref}
\usepackage{amssymb}
\usepackage{amsfonts}
\usepackage{bm} 
\usepackage{xspace}
\usepackage{booktabs}
\usepackage{multirow}
\usepackage{caption}
\usepackage{subcaption}
\usepackage{url}

\newcommand\blfootnote[1]{%
  \begingroup
  \renewcommand\thefootnote{}\footnote{#1}%
  \addtocounter{footnote}{-1}%
  \endgroup
}

\input{macros}
\begin{document}
\title{FairGLVQ: Fairness in \\ Partition-Based Classification\footnote{This preprint has
not undergone any post-submission improvements or corrections. The Version of Record of this contribution is published in Advances in Self-Organizing Maps, Learning Vector Quantization, Interpretable Machine Learning, and Beyond,
and is available online at \url{https://doi.org/10.1007/978-3-031-67159-3_17}}}

\author{Felix Störck${}^{\dagger}$, Fabian Hinder, Johannes Brinkrolf, \\ Benjamin Paassen, Valerie Vaquet, Barbara Hammer \\\;\\
\small Bielefeld University \\
\small Inspiration 1, 33619 Bielefeld  - Germany\\
\small \texttt{\{fstoerck, fhinder, jbrinkro, bpaassen, \qquad\qquad} \\ \small \texttt{\qquad\qquad vvaquet, bhammer\}@techfak.uni-bielefeld.de}}
\maketitle             
\blfootnote{\!\!\!\!${}^\dagger$ Corresponding Author}
\begin{abstract}
Fairness is an important objective throughout society. From the distribution of limited goods such as education, over hiring and payment, to taxes, legislation, and jurisprudence. Due to the increasing importance of machine learning approaches in all areas of daily life including those related to health, security, and equity, an increasing amount of research focuses on fair machine learning. 
In this work, we focus on the fairness of partition- and prototype-based models. 
The contribution of this work is twofold: 1) we develop a general framework for fair machine learning of partition-based models that does not depend on a specific fairness definition, and 2) we derive a fair version of learning vector quantization (LVQ) as a specific instantiation. We compare the resulting algorithm against other algorithms from the literature on theoretical and real-world data showing its practical relevance. 

\textbf{keywords: }{Fair Machine Learning $\cdot$ Fairness $\cdot$ Partition-based Models $\cdot$ Prototype-based Models $\cdot$ Fair LVQ.}
\end{abstract}

\section{Introduction}

Whenever the outputs of classification models impact human lives, fairness is a crucial concern. For example, if models suggest grades in school, score applications in human resources, or distribute scarce resources like in water networks, individuals and groups have good reason to demand equal treatment \cite{mehrabi_survey_2021,Strotherm2023,Strotherm2023b}. However, fairness alone is insufficient for the requirements of modern socio-technical systems. Especially in the European context, models also need to facilitate explainability and transparency to be eligible for the label \enquote{trustworthy artificial intelligence} as introduced by the EU \cite{HLEG2019} and as codified in the European AI act \cite{Laux2023}. Further requirements may be technical in nature, such as compatibility with streaming data or data sparsity.

While extensive prior work on fairness in machine learning exists, most of it has focused solely on the relation of fairness and accuracy \cite{mehrabi_survey_2021,Strotherm2023b}, whereas other requirements have been mostly ignored. Prior efforts to make interpretable models more fair have been almost exclusively focused on decision trees \cite{kamiran_discrimination_2010,van_der_linden_fair_2022,ranzato_fairness-aware_2021}.

In this paper, we generalize beyond this prior stream of work by providing a general framework for fairness in partition-based models, a class of models that includes (but is not limited to) decision trees. We also provide a new implementation of fair partition-based machine learning for learning vector quantization (LVQ) models, thus demonstrating the generality of our proposed framework and providing a novel algorithm for fair machine learning that is intrinsically interpretable and compatible with strict resource constraints (constant memory, streaming data, data scarcity). The resulting approach allows for a local and non-linear adaption to fairness constraints. In particular, our method is able to outperform methods based on feature selection and iterative null-space projections which are important state-of-the-art methods in fair machine learning but in the case of composed datasets \cite{IRMA,ravfogel-etal-2020-null}.

This paper is structured as follows: First (\cref{sc:setup}), we recall the fundamental setup of fairness, focusing on group fairness. We then recall the structure of LVQ models (\cref{sc:lvq}) and briefly discuss the related work on fair machine learning models (\cref{sc:rel_work}). Afterward, we introduce our framework for fair partition-based models (\cref{sc:general_method}), from which we derive a fair version of LVQ (\cref{sc:flvq}) by viewing it through the lens of Hebbian learning (\cref{sc:hebbian}). We empirically evaluate our method (\cref{sc:exp}) on synthetic and real-world data and conclude this work (\cref{sc:conclusion}).

\section{Problem Setup}
\label{sc:setup}

In this section, we recall the setup of fair machine learning focusing on group fairness as well as the vanilla LVQ models and give a short overview of the related work.

\subsection{Fair Machine Learning}
\label{sc:fml}
In supervised machine learning, one commonly considers models $\clf{} : \domain \to \codomain$ that map samples $\point$ in the data space $\domain$ to a target or label $\outpoint$ in the target space $\codomain$. 
Evaluating a model is done by applying a loss function $\loss : \codomain \times \codomain \to \R$, e.g. 0-1-loss for classification, MSE for regression, etc., to compare the model output to the desired target $\Loss_\dataset(\clf{}) = \frac{1}{n} \sum_{i = 1}^\datalim \loss(\clf{\point_\dataidx},\outpoint_\dataidx)$ for a data set $\dataset =\lbrace(\point_\dataidx,\outpoint_\dataidx)\in \domain\times\codomain \mid\dataidx=1,\dots, \datalim\rbrace$. Training is usually done by optimizing the model parameters such that the loss is minimized.
In general, this procedure can lead to models that exhibit undesired biases or unexpected drastic changes in the output which are considered as unfair. 

In fair machine learning, the training objective is modified to cope with such problems. Depending on the specific notion of fairness, different objectives are used:
\emph{Individual fairness} requires that similar data points are assigned similar outputs. Here, the notion of similarity is usually hand-defined by a human operator based on predefined rules~\cite{mehrabi_survey_2021}.

In this work, we will mainly focus on \emph{group fairness} which extends the label data pair by a \emph{protected} or \emph{sensitive attribute} $(\point,\outpoint,\protpoint)$, $\protpoint \in \protdomain$, e.g., sex, race, age, etc. In this case, the training objective is modified to reduce biases that can be linked to the protected attribute which can be evaluated using different scores:
\emph{Statistical Parity} describes the idea that all protected groups should have the same amount of \textit{favorable} decisions \cite{mehrabi_survey_2021}. 
\emph{Equal Opportunity} on the other hand focuses only on the cases where the true label is favorable as well \cite{mehrabi_survey_2021}.

In general, fairness and accuracy cannot be expected to both be optimal, a phenomenon referred to as \emph{fairness-accuracy-tradeoff}. Moreover, different measures are incompatible with each other \cite{mehrabi_survey_2021} and some can be full-filled in discriminatory settings \cite{Dwork_2012}.
This leads to the problem of choosing an appropriate measure of fairness.

\subsection{Learning Vector Quantization Models}
\label{sc:lvq}
In this contribution, we focus on partition-based classifiers, particularly, on Learning Vector Quantization, as intuitive and distance-based classifiers which are designed to solve the multi-classification problem for real vectorial data, i.e., $\domain \subset \R^\datadim$ and $\codomain = \{1,\dots,\numClasses\}$.
They are defined as follows: prototypes $\proto_1,\dots,\proto_\protolim$ with fixed labels $\class{\proto_\protoidx}\in\codomain$ are specified
such that a good classification and representation of the data is achieved. A new sample $\point$ is classified by the winner-takes-all scheme:
\begin{equation}
	\label{eq:back_winnerTakesAll}
\point\mapsto \class{\proto_{J(\point)}}\ \mbox{ where }\  J(\point):=\argmin_{\protoidx\in\lbrace1,\dots, \protolim\rbrace} \dis{\point}{\proto_\protoidx},
\end{equation}
where 
$\dis{\point}{\proto_\protoidx}=\transp{(\point-\proto_\protoidx)}(\point-\proto_\protoidx)$
denotes the squared Euclidean metric.

Given a data set $\dataset$, 
prototypes $\proto_\protoidx$ are adapted to minimize the classification error. 
In this work, we focus on GLVQ which is based on margin maximization~\cite{sato_generalized_1995,schneider_adaptive_2009}. In this case, the cost function is given by
\begin{equation}
\label{eq:glvq_costfct}
\errGLVQ=\sum_{\dataidx=1}^\datalim \nonlin\left(\mu(\point_\dataidx)\right),
\end{equation}

\begin{equation}
\label{eq:glvq_reldist}
\mu(\point_\dataidx)=\frac{\dis{\point_\dataidx}{\protop}-\dis{\point_\dataidx}{\protom}}{\dis{\point_\dataidx}{\protop}+\dis{\point_\dataidx}{\protom}}
\end{equation}
where $\nonlin$ is a monotonic increasing function, \eg the logistic function or the identity $ \nonlin(\point)=\point $,
$ \protop $ denotes the prototype closest to $\point_\dataidx$ belonging to the same class as $\point_\dataidx$, i.e., $\class{\protop} = \outpoint_\dataidx$, and the closest prototype $ \protom$ of a different class, i.e., $\class{\protom} \neq \outpoint_\dataidx$. 
The term $\mu(\point_\dataidx)\in [-1,1] $ 
is the relative distance difference between the closest correct and incorrect prototype.
Depending on the sign a point is either correctly ($\mu(\point_\dataidx)<0$), incorrectly classified ($>0$), or on the decision boundary ($=0$).

Training takes place based on a given training set as above and consists of two steps. First, the prototypes are initialized, typically this is either done within the class centers or by using the cluster centers of a clustering algorithm. 
Second, the cost function $\errGLVQ$ is minimized using a gradient descent scheme.
In this work, we choose $\nonlin$ as the swish activation function \cite{villmann_2020}.

\subsection{Related Work}
\label{sc:rel_work}
There is a rich body of literature on fair machine learning.
According to \cite{mehrabi_survey_2021}, creating fair models can be split into three main segments:
1) \emph{pre-processing} the dataset reduces biases before training, 2) controlling the bias during training of the model using a suited objective (\emph{in-processing}), and 3) \emph{post-processing} the trained model to remove potentially learned biases.

Common pre-processing approaches transform the data by removing the protected attribute or projecting onto an orthogonal subspace (\emph{null-space}) \cite{ravfogel-etal-2020-null}. This process can be iterated multiple times. We refer to this method as ``iterative null-space projection'' (INP). However, such methods can remove valuable information while information on the protected attribute can still be contained in (highly non-linear) correlations~\cite{ravfogel-etal-2020-null,mehrabi_survey_2021}.

A simple in-processing technique is to include the desired fairness criterion as a loss term or constraint during training \cite{zhang_faht_2019,Kamishima_2012}.

Adversarial Debiasing~\cite{zhang_mitigating_2018} is a state-of-the-art in-process technique that uses ideas similar to \cite{ravfogel-etal-2020-null}: A deep representation that is well suited for the learning problem but unsuited for predicting the protected attributed is learned using a kind of adversarial training. A disadvantage of this approach is the instability of adversarial learning~\cite{zhang_mitigating_2018}.

Several approaches exist to train fair decision trees~\cite{kamiran_discrimination_2010,van_der_linden_fair_2022,ranzato_fairness-aware_2021}.
A particular simple in-processing technique is proposed by \cite{kamiran_discrimination_2010}. Here, the decrease in impurity of the protected attribute is considered as a penalty in the splitting criterion.

Although rich literature exists on partition-based models, to the best of our knowledge, there are no previous works on the fairness of LVQ models.

\section{Method}
In this section, we first present a general method for training fair partition-based models. 
We then apply this scheme to GLVQ to derive a novel, fair LVQ variant. 

\subsection{Fairness for General Partition-Based Models} \label{sc:general_method}
The general concept of partition-based models relies on partitioning the data space into regions that are homogeneous w.r.t to the targets. Each region is targeted using a single, usually simple model. We will focus on constant models, the resulting partition-based models are thus given by $\clf{\point} = \clsop\circ\partition(\point)$ where $\partition : \domain \to \N$ is a partition and $\clsop : \N \to \codomain$. Notice that both decision trees as well as LVQ belong to this type of model. In the case of decision trees, the regions are given by the leaves; in the case of LVQ, they are induced by the prototypes. 

Given a dataset, $\dataset$, partition-based models can be trained by minimizing the loss over $\partition$ and $\clsop$ as usual. 
While optimizing $\partition$ may be difficult, optimizing $\clsop$ given $\partition$ is usually simple, e.g.\ the average in the partition for decision trees.

The main problem is thus to find the right partition so that the regions are homogeneous with respect to the target. To ensure fairness, we use the opposite criterion for the protected attribute, i.e., we reward a partition for a low loss on predicting the targets (\emph{local homogeneity}) but also for a high loss for predicting the protected attribute (\emph{local heterogeneity}). We achieve this by adding a regularization term, weighted by a scaling constant $C$, yielding a new optimization problem:
\begin{equation}
\label{eq:partitionmodel}
\min_{\partition} \left[\min_{\clsop} \Loss^\text{class}_\dataset(\clsop\circ\partition) - C \cdot \min_{\fairclsop} \Loss^\text{fair}_\dataset(\fairclsop\circ\partition)\right],
\end{equation}
where $\Loss^\text{class}_\dataset$ and $\Loss^\text{fair}_\dataset$ are the empirical losses concerning the target and protected attribute, respectively.
We emphasize that this formulation is \emph{independent of a specific fairness metric} as we generally punish good classification w.r.t.\ the protected attribute. It also
directly \emph{applies to regression} for both target and protected attributes.

In the next step, we discuss our training scheme from a Hebbian perspective before applying it to LVQ.

\subsection{Fair Hebbian Learning, Pseudo-Classes, and FairGLVQ}
\label{sc:hebbian}
The fundamental idea of Hebbian learning is to reinforce or weaken the connection between neurons that are active at the same time. In the case of classical LVQ, this is realized by moving the prototypes closer to or further away from the observed data point. To assure fairness similar to \cref{eq:partitionmodel}, we also want to have an inhibiting effect if a neuron is strongly associated with the protected attribute. For LVQ, this means pushing the prototype away.

This effect can easily be obtained by assigning a second \emph{pseudo-class} representing the majority protected attribute (cf. \cref{eq:partitionmodel}) to each prototype that is used with an opposite gradient during training. Notice that the pseudo-class is updated during the training process.
We will discuss more technical details in the next section where we give an explicit algorithm as an extension of GLVQ: FairGLVQ. 

\subsection{Algorithm Design}
\label{sc:flvq}
This section describes the technical details of how our algorithm is designed to implement the general method presented in \cref{sc:general_method}.
Following the ideas from the last section, we adapt \cref{eq:glvq_costfct} to punish good classification w.r.t.\ the pseudo-classes.
Let $\mu_\text{fair}(\point_\dataidx)$ denote \cref{eq:glvq_reldist} for the pseudo-classes. Then, the new FairGLVQ cost function is given as:

\begin{equation}
    \label{eq:fglvq_loss}
    E_\text{fair} = \sum_{i=1}^n \nonlin(\mu_\text{class}(\point_\dataidx))-C\cdot\nonlin(\mu_\text{fair}(\point_\dataidx)).
\end{equation}
The parameter $C$ again controls the strength of the regularization.
The resulting algorithm is presented in \cref{alg:FGLVQ}. Observe that up to lines \ref{algo:FGLVQ:start_fair} to \ref{algo:FGLVQ:end_fair} this is just a mini-batch version of the standard GLVQ training algorithm. Here we use a mini-batch approach as the computation of the fair loss requires the pseudo-classes which are expensive to compute.

As the pseudo-classes are not fixed but reassigned according to the majority vote, we have to take two edge cases into account. 

\emph{Edge-Case 1: empty receptive field:} we choose a pseudo-class at random.

\emph{Edge-Case 2: no (in-)correct pseudo-class:} computing $\mu_\text{fair}$ requires the closest prototype with the same/different pseudo-class which may not exist. 
In this case, we define $\dis{\point}{\protom_\text{fair}}\coloneqq \frac{1}{\alpha}\cdot\dis{\point}{\protop_\text{fair}}$ and  $\dis{\point}{\protop_\text{fair}}\coloneqq \alpha\cdot\dis{\point}{\protom_\text{fair}}$ if $\protom_\text{fair}$ or $\protop_\text{fair}$ does not exist, respectively, which corresponds to simulating a prototype such that $\point$ is considered to have a protected attribute different from the overall closest prototype (lines 8-14 in \cref{alg:FGLVQ}).
This setting implies $\mu_\text{fair}(\point_\dataidx)=\frac{\alpha - 1}{\alpha + 1}\in(0,1)$.
Therefore, the parameter $\alpha\in(1,\infty)$ determines 
the value of the fairness penalty in this case. 
We use $\alpha=2$ as a default value. 

\begin{algorithm}[t]
   \caption{FairGLVQ}
   \label{alg:FGLVQ}
\begin{algorithmic}[1]
   \State {\bfseries Input:} {$\dataset$ train data, $N$ number of epochs, $M$ batch-size, $\eta$ learning rate}, regularization parameter $C$, missing pseudo-class parameter $\alpha$
   \State Initialize prototypes $\vec{W} \gets \{\proto_1, ..., \proto_\protolim\}$  \Comment{Use fairness-adapted clustering}
   
   \For{$N$ steps} 
   \State $\mathbf{G}^{\text{class}} \gets (0,\dots,0)$, $\mathbf{G}^{\text{fair}} \gets (0,\dots,0)$, $n_{\text{fair}} \gets 0$, $n_{\text{class}} \gets 0$
   \State Pick mini-batch $B=\{(\point_1,\outpoint_1,\protpoint_1),(\point_2,\outpoint_2,\protpoint_2),...,(\point_M,\outpoint_M,\protpoint_M) \} \subset \dataset$  
       \ForAll{$(\point,\outpoint, \protpoint)\in B$}
        \State $\protopm_{\text{class}\atop\text{fair}} \gets \argmin_{{\proto{}} \in \vec{W}^\pm_{\text{class}\atop\text{fair}}} \dis{\point}{\proto{}}$ 
        \Comment{Determine $\protop_\text{class},\protom_\text{class},\protop_\text{fair},\protom_\text{fair}$}
        \If{$\protom_\text{fair}$ does not exist} \label{algo:FGLVQ:start_fair}
            \State $\dis{\point_\dataidx}{\protom_\text{fair}}\coloneqq \frac{1}{\alpha}\cdot\dis{\point_\dataidx}{\protop_\text{fair}}$, $\vec{W}_\text{update}\gets\{\protop_\text{fair},\protop_{\text{class}},\protom_{\text{class}}\}$
        \ElsIf{$\protop_\text{fair}$ does not exist}
            \State $\dis{\point_\dataidx}{\protop_\text{fair}}\coloneqq \alpha\cdot\dis{\point_\dataidx}{\protom_\text{fair}}$, $\vec{W}_\text{update}\gets\{\protom_\text{fair},\protop_{\text{class}},\protom_{\text{class}}\}$
        \Else  
            \State $\vec{W}_\text{update}\gets\{\protop_\text{fair},\protom_\text{fair},\protop_{\text{class}},\protom_{\text{class}}\}$
        \EndIf \label{algo:FGLVQ:end_fair}

        \State $\mathbf{G}^{\text{class}\atop\text{fair}}_{\proto{}} \gets \mathbf{G}^{\text{class}\atop\text{fair}}_{\proto{}} + \nabla_{\proto{}} E_\text{fair}$ for $\proto{} \in \vec{W}_\text{update} $
        \Comment{Compute gradient}
        \State $n_\text{fair} \gets n_\text{fair} + |\vec{W}_\text{update}|-2$ , $n_\text{class} \gets n_\text{class} + 2$
        
       \EndFor
       \ForAll{$\proto \in \vec{W}$} \Comment{Update prototypes using mini-batch gradients $\mathbf{G}^{\text{class}\atop\text{fair}}_{\proto{}}$}
            \State ${\proto{}} \gets {\proto{}} - \eta\left(\frac{1}{n_{\text{class}}} \mathbf{G}^{\text{class}}_{\proto{}} + \frac{1}{n_{\text{fair}}} \mathbf{G}^{\text{fair}}_{\proto{}}\right)$  
       \EndFor
       \ForAll{$\proto_{j} \in \vec{W}$} \Comment{Update pseudo-classes using majority vote}
            \State $\protpoint({\proto_{j}}) \gets \argmax_{\protpoint \in \protdomain} |\{ (\point,\outpoint, \protpoint) \in \dataset \mid \protpoint = \protpoint({\proto_{j}}) \wedge J(\point) = j \}|$ 
            
       \EndFor
\EndFor
\end{algorithmic}
\end{algorithm}

In contrast to the usual LVQ initialization which is class-specific, we initialize prototypes for every class at every found cluster center with a small perturbation. 

\section{Experiments}
\label{sc:exp}
In this section, we empirically evaluate our approach comparing it to other methods from the literature.\footnote{The code and datasets can be found at \url{https://github.com/Felix-St/FairGLVQ}} We consider both, a synthetic dataset to showcase the specific properties of FairGLVQ as well as 
two real-world datasets to demonstrate the practical relevance of our approach.

For both experiments, we make use of three baselines: GLVQ, GLVQ with linear pre-processing (INP), and a constant baseline model.
For GLVQ, we initialize the prototypes based on a per-class clustering using $k$-means, otherwise with the same parameters as we use for FairGLVQ. For the linear pre-processing, we use the iterative null-space projection method (INP) \cite{ravfogel-etal-2020-null} and apply standard GLVQ to the pre-processed data. Notice that for $C=0$ or no projection FairGLVQ and INP are equal to GLVQ (up to prototype initialization).

\subsection{Synthetic Data}
\begin{figure}[t]
    \centering
	\includegraphics[width=0.7\textwidth]{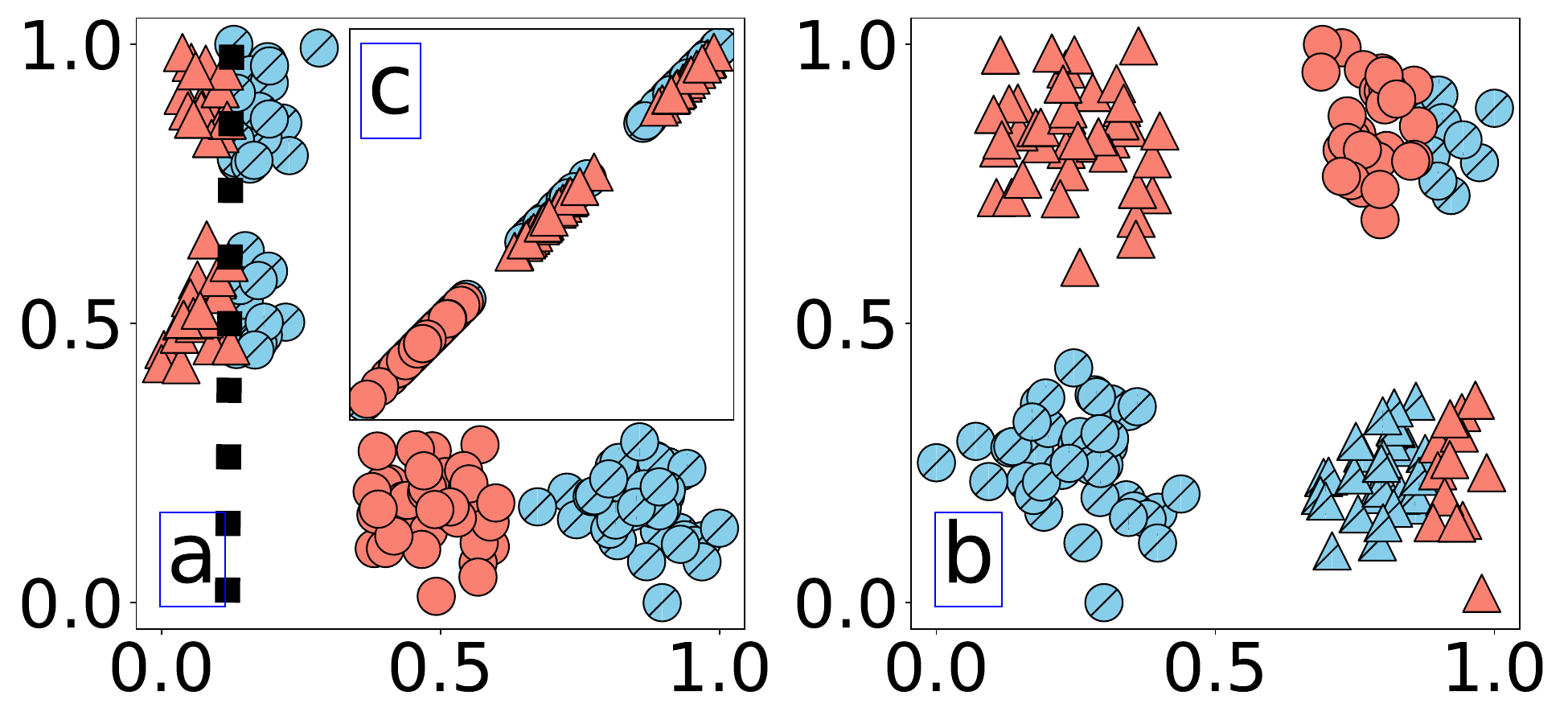}
    \caption{Synthetic datasets (\emph{a}: ``local'' with linear model (black), \emph{b}: ``XOR'', \emph{c}:~fair projection of ``local''). Label is color/hatched; Protected attribute is shape.}
	\label{fig:Synthetic_Datasets}
\end{figure}

We start by showcasing the advantages of FairGLVQ on synthetic data. We consider two datasets with 2 classes and a binary protected attribute (See \cref{fig:Synthetic_Datasets} for a visualization). ``XOR'' consists of 4 Gaussian's, the occurrence of classes and protected values can be described by xors shifted against each other along one axis. ``local'' consists of two separate classification problems, one aligned with the protected attribute, the other unaffected by the protected attribute. The protected attribute is not considered as part of the data.

\begin{table}[t]
\caption{ Mean (and standard deviation) Accuracy (Acc; higher is better) and Statistical Parity (SP; lower is better) on synthetic datasets over a 5-fold cross-validation.} 
    \centering
    \begin{tabular}{l@{\:\:}lcccc}
\toprule
 && $\underset{\text{(ours)}}{\text{FairGLVQ}}$ & INP & GLVQ & const.  \\
\midrule \multirow{2}{*}{\rotatebox[origin=c]{90}{Local}}
& Acc &$0.76(\pm 0.02)$ & $0.51 (\pm 0.01)$ & $0.99(\pm 0.00)$ & $0.51(\pm 0.00)$\\
& SP &$0.12(\pm 0.13)$ & $0.24 (\pm 0.12)$ & $0.65(\pm 0.01)$ &  $0.00(\pm 0.00)$\\
\midrule \multirow{2}{*}{\rotatebox[origin=c]{90}{XOR}}
& Acc &$0.88(\pm 0.01)$ & $0.84 (\pm 0.05)$ & $0.99(\pm 0.00)$ & $0.50(\pm 0.00)$\\
& SP &$0.03(\pm 0.02)$ & $0.24 (\pm 0.12)$ & $0.23(\pm 0.04)$ &  $0.00(\pm 0.00)$\\
\bottomrule
\end{tabular}
    \label{tb:synthetic_results}
\end{table}

We use $4$ prototypes per class for XOR and $5$ for local. 
As the dataset is two-dimensional we project onto one dimension for INP. For FairGLVQ, we use $M=250$, $\alpha=2$, $\eta=0.005$, and $C=1.5$ for local and $C=1.25$ for XOR. We train for $N=250$ epochs.
We run a 5-fold cross-validation and measure Accuracy and Statistical Parity (the latter as the absolute difference). The results are shown in \cref{tb:synthetic_results}.
As can be seen, GLVQ has the highest accuracy but is very unfair. FairGLVQ outperforms INP with regard to accuracy and shows comparable fairness. For the ''local'' dataset, this can be explained by observing that there is no fair projection that still provides sufficient information for classification as illustrated in \cref{fig:Synthetic_Datasets}. 
For the ''XOR'' dataset, a linear classifier cannot reliably learn a useful projection dimension.
This shows the advantage of the local and non-linear fairness criterion considered by FairGLVQ.

\subsection{Real-world Benchmarks}
Besides theoretical datasets, we consider the real-world datasets ``COMPAS''~\cite{julia_angwin_machine_nodate} (protected attribute race) and ``Adult Income''~\cite{misc_census_income_20} (Adult; protected attribute sex). In both cases, the protected attribute is provided as a feature.

For both datasets, we used $20$ prototypes per class, and $500$ epochs with a learning rate of $0.05$. For FairGLVQ, we used $\alpha=2$, $M=200$ for COMPAS, and $M=1000$ for Adult. We considered different regularization strengths, i.e., $C$ for FairGLVQ and the number of projection dimensions for INP.

\begin{figure}[t]
	\centering
    \begin{minipage}{\textwidth}
        \centering
        \includegraphics[width=1\textwidth]{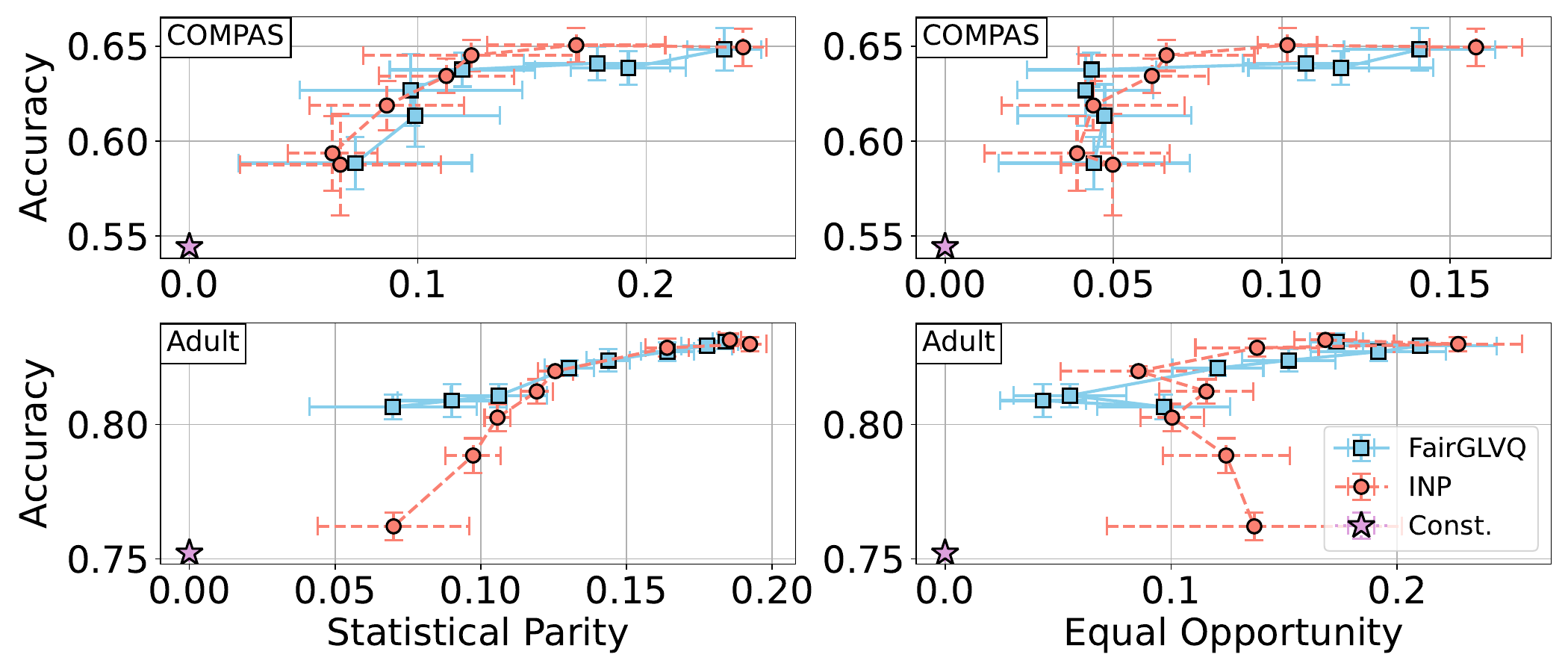}
    \end{minipage}
    \caption{Empirical evaluation of real-world datasets for different regularizations using 5-fold cross-validation. Fairness score ($x$-axis, lower is better) and Accuracy ($y$-axis, higher is better). COMPAS (top) and Adult (bottom).}
	
	\label{fig:Eval}
\end{figure}

We document Accuracy, Statistical Parity, and Equal Opportunity (the latter two as absolute difference) using a 5-fold cross-validation. The results are shown in \cref{fig:Eval}.
As shown, both algorithms are finding good compromises within the fairness-accuracy-tradeoff for the COMPAS dataset and perform very similarly with slight differences in high and low regularization tradeoffs respectively.
For Adult, INP is only able to improve fairness up to a certain point before Accuracy drops to a constant baseline level.
One explanation for this effect is that the projection removes so much information that classification is no longer possible, similar to the ``local'' data set. 
In particular, FairGLVQ outperforms INP on Adult in the case of high fairness. 
Conversely, this implies that COMPAS admits a linear decomposition into subspaces relevant for predicting the label and protected attribute, respectively.

Note that both approaches improve in both fairness metrics. However, for the Adult dataset, the regularization does not necessarily lead to a monotonic decline for weak regularization in both algorithms and for strong regularization in FairGLVQ. Otherwise, the relation is monotonic as expected.

\section{Conclusion and Discussion}
\label{sc:conclusion}
In this paper, we considered the problem of group fairness for partition-based models with a particular focus on LVQ. We derived a general framework to impose fairness in partition-based models via an intuitive notion of fairness (punishing good classification w.r.t.\ protected attribute) that is not limited to discrete protected attributes nor specific fairness metrics like Statistical Parity or Equal Opportunity. 

Based thereon, we derived a new, fair version of the classical LVQ models by considering the framework through the lens of Hebbian learning. Therefore, we believe our considerations can also be applied to other models and methods. In particular, the application to other LVQ models like GMLVQ \cite{schneider_adaptive_2009}, LGMLVQ \cite{schneider_adaptive_2009} or KGLVQ \cite{Qin_2004} is subject to further work. 

We evaluated FairGLVQ, showing a good fairness-accuracy-tradeoff. In particular, we showed that it can outperform global approaches based on iterated null-space projection, a state-of-the-art method for fairness and federated learning. Probing the potential of FairGLVQ for handling composed datasets is yet another interesting line of further work as well as investigating the performance on multi-class / multi-protected problems. Due to the locality of our approach, an investigation into individual fairness is another interesting frontier.

\subsubsection*{Acknowledgement} We gratefully acknowledge funding from the European Research Council (ERC) under the ERC Synergy Grant Water-Futures (Grant agreement No. 951424).

\bibliographystyle{plain}
\footnotesize
\bibliography{literature}
\end{document}

%% file: macros.tex
\newcommand*{\eg}{e.\,g.\@\xspace}

\newcommand*{\transp}[1]{{#1^{\top}}\allowbreak }

\newcommand*{\R}{\mathbb{R}}

\newcommand*{\N}{\mathbb{N}}
\newcommand{\domain}{\ensuremath{\mathcal{X}}}
\newcommand{\codomain}{\ensuremath{\mathcal{Y}}}
\newcommand{\protdomain}{\ensuremath{\mathcal{S}}}
\newcommand{\dataset}{\mathcal{D}}

\renewcommand{\vec}[1]{\bm{#1}}
\newcommand{\point}{\vec{x}}
\newcommand{\outpoint}{y}
\newcommand{\protpoint}{s}
\newcommand{\proto}{\vec{w}}
\newcommand{\protoidx}{j}
\newcommand{\protolim}{P}
\newcommand{\errGLVQ}{E}
\newcommand{\nonlin}{\Phi}

\newcommand{\signp}{+}
\newcommand{\signm}{-}
\newcommand{\protop}{\proto^\signp}
\newcommand{\protom}{\proto^\signm}
\newcommand{\protopm}{\proto^\pm}

\DeclareMathOperator*{\clfop}{h}
\DeclareMathOperator*{\clsop}{c}
\DeclareMathOperator*{\partition}{p}
\DeclareMathOperator*{\fairclsop}{f}
\newcommand{\clf}[1]{\clfop\ifthenelse{\equal{#1}{}}{}{( #1 )}}
\newcommand{\class}[1]{\clsop( #1 )}
\newcommand{\numClasses}{c}

\newcommand{\datadim}{d}
\newcommand{\dataidx}{i}
\newcommand{\datalim}{n}

\newcommand{\loss}{\ell}
\newcommand{\Loss}{\mathcal{L}}

\newcommand{\dischar}{d}
\newcommand{\dis}[3][]{\dischar\ifthenelse{\equal{#1}{}}{}{_{#1}} \ifthenelse{\equal{#2}{}}{}{(#2,\,#3)}}

\DeclareMathOperator*{\argmax}{arg\,max}
\DeclareMathOperator*{\argmin}{arg\,min}